\documentclass[journal]{IEEEtran}
\IEEEoverridecommandlockouts
\usepackage{cite}
\usepackage{amsmath,amssymb,amsfonts}
\usepackage{graphicx}
\usepackage{subcaption}
\usepackage{hyperref}
\usepackage{color}
\usepackage{tabularx}
\usepackage{soul}
\PassOptionsToPackage{hyphens}{url}\usepackage{hyperref}
\usepackage{cleveref}
\usepackage[linesnumbered,ruled,vlined]{algorithm2e}
\usepackage{mdframed}

\renewcommand{\baselinestretch}{0.984}
\begin{document}

\title{Multiverse of Greatness: Generating Story Branches with LLMs}

\author{Pittawat Taveekitworachai, Chollakorn Nimpattanavong, Mustafa Can Gursesli,\\ Antonio Lanata, Andrea Guazzini, and Ruck Thawonmas%
    \thanks{P. Taveekitworachai and C. Nimpattanavong are with the Graduate School of Information Science and Engineering, Ritsumeikan University, Osaka, Japan. Emails: research@petepittawat.dev and gr0608sp@ed.ritsumei.ac.jp}%
    \thanks{M. C. Gursesli and A. Lanata, are with the Department of Information Engineering, University of Florence, Florence, Italy. Emails: mustafacan.gursesli@unifi.it and antonio.lanata@unifi.it}%
    \thanks{A. Guazzini is with the Department of Education, Literatures, Intercultural Studies, Languages and Psychology, University of Florence, Florence, Italy. Email: andrea.guazzini@unifi.it}%
    \thanks{R. Thawonmas is with the College of Information Science and Engineering, Ritsumeikan University, Osaka, Japan. Email: ruck@is.ritsumei.ac.jp}%
}

\markboth{Preprint}%
{Taveekitworachai \MakeLowercase{\textit{et al.}}: Multiverse of Greatness: Generating Story Branches with LLMs}

\maketitle

\begin{abstract}
This paper presents Dynamic Context Prompting/Programming (DCP/P), a novel framework for interacting with LLMs to generate graph-based content with a dynamic context window history. While there is an existing study utilizing LLMs to generate a visual novel game, the previous study involved a manual process of output extraction and did not provide flexibility in generating a longer, coherent story. We evaluate DCP/P against our baseline, which does not provide context history to an LLM and only relies on the initial story data. Through objective evaluation, we show that simply providing the LLM with a summary leads to a subpar story compared to additionally providing the LLM with the proper context of the story. We also provide an extensive qualitative analysis and discussion. We qualitatively examine the quality of the objectively best-performing generated game from each approach. In addition, we examine biases in word choices and word sentiment of the generated content. We find a consistent observation with previous studies that LLMs are biased towards certain words, even with a different LLM family. Finally, we provide a comprehensive discussion on opportunities for future studies.
\end{abstract}


\section{Introduction}\label{sec:intro}

A visual novel game has unique characteristics in offering freedom to players in making choices that potentially affect the course of the storyline and its ending \cite{finley2022branching}. It is obvious that storylines are a core part of the visual novel experience and what sets it apart from other game genres. This unique characteristic of choice opportunities throughout the story allows the player to interact and choose to branch into a new storyline. We can view a visual novel game as a tree traversal, starting from a root and traversing down the tree based on choices made throughout the storyline, forming a branch and reaching a leaf node at the end of the story. More strictly, visual novel storylines are often represented by a directed acyclic graph (DAG), as there is a possibility that each player may have a different starting node and it is possible for two storylines to merge into the same storyline at some point.

Despite this fundamental characteristic of visual novels, we found that an existing study \cite{gursesli2023chronicles} utilizing large language models (LLMs) \cite{zhao2023survey} for generating a visual novel game employed a very basic approach for generating a short visual novel game. Not only did their approach not take into account this fundamental property as a graph of visual novel stories, but it also required extensive human labor in extracting the LLM's responses and crafting a game. On the other hand, there already exist studies utilizing LLMs to generate narrative elements compatible with graph-based data \cite{riedl2006from}, showing us the potential for utilizing LLMs to fully generate content as a graph and automatically generate a visual novel game.

Nevertheless, utilizing LLMs to generate a visual novel game is not trivial. LLMs have a limited context window \cite{chen2023extending}, and visual novel games present a challenge in fitting all story branches into a limited context window. Moreover, we discovered that simply prompting LLMs with an initial story data without giving the previous context of the actual story branch does not work well. Therefore, we propose a novel approach in utilizing LLMs to generate graph-based content with a dynamic context window, named \textbf{dynamic context prompting/programming} (DCP/P). We leverage an idea that has already existed in the computer science algorithm community for a long time in solving problems with graphs, called dynamic programming, and common graph traversal algorithms, such as breadth-first search, to craft our novel approach in interacting with LLMs. Furthermore, we also suggest that we can view LLMs as a magical, arbitrary, text-generation function that can be programmed via prompting.

We also note that our approach lets LLMs decide almost everything, e.g., who the main characters are, what possible locations are, what possible choices are, how each choice is going to affect the course of the story, and when to use a narrator or a character to progress the story. There is minimal interference from the user of our framework, and most decisions are made by LLMs. However, we also accept a number of arguments from the user to offer control over what type of game is generated and the approximate length of the game; the framework accepts themes that the user wants to focus on for the game, how many chapters exist in the game, what the minimum and maximum number of choices at each opportunity are, and more.

A visual novel game does not only contain text but also images, hence the word visual in the name. In order to make our approach able to generate a visual novel game, and not simply a text-based adventure game, we also incorporate a mechanism to utilize an image generative model \cite{zhang2023texttoimage} to generate game assets for each game. While audio is another important part of the game experience, we opted not to pursue it in this current study and plan to explore this aspect in the future.

To evaluate our DCP/P, we compare it against a baseline of simply prompting LLMs using the same prompts and framework, but without providing previous context and only relying on the initial story data. We follow an existing study \cite{gursesli2023chronicles} to objectively evaluate the linguistic aspects of the generated stories. We generate a total of 10 stories for each approach, baseline and DCP/P, resulting in a total of 20 stories. We observe that games generated using DCP/P consistently outperforms the baseline's games. To further understand the generated stories, we provide a qualitative analysis and discussion for the best-performing game based on the objective score from each approach. Moreover, an existing study \cite{taveekitworachai2023breaking} suggested that LLMs may be biased towards certain words or specific types of sentiment; we also evaluate these aspects following an existing approach.

We open-source all of our code and prompts utilized for this framework to support further research. This also includes our custom web user interface that can display any arbitrary game generated using our approach without needing to change any code. We also make it possible to extend our framework and utilize any LLMs or image generative models available with minimal changes for greater possibility. Therefore, our contributions are as follows:

\begin{itemize}
    \item We propose a novel approach in interacting with LLMs for generating a visual novel game, DCP/P.
    \item We extensively perform objective evaluation and comprehensive discussions of the generated games against the baseline approach.
    \item We open-source all of our research artifacts\footnote{\begin{itemize}
        \item Algorithm: \url{https://github.com/multiverse-of-greatness/story-branching-algorithms}
        \item UI: \url{https://github.com/multiverse-of-greatness/multiverse-ui}
        \item Utility code: \url{https://github.com/multiverse-of-greatness/story-branching-utils}
        \item Evaluation algorithm: \url{https://github.com/multiverse-of-greatness/evaluation-algorithm}
    \end{itemize}} to support future research.
\end{itemize}

\section{Related Work}\label{sec:related_work}
\subsection{LLMs for Narrative Generation}\label{sec:related_work:narrative}

In video games, various factors such as graphics quality, physics mechanics, and in-game audio contribute considerably to the player's immersion and enjoyment \cite{gallacher2013game,cheng2005behaviour}. Moreover, the depth and sophistication of the game's narrative, including its characters and dialogue, are critical in creating a compelling and engaging experience \cite{naul2020story}. A well-written narrative can enhance the player's connection to the game, making narrative design a crucial component of game development \cite{ermi2005fundamental}.

Creating complex and diverse character-driven stories has traditionally been a challenging task for human writers. However, advances in LLMs have provided significant opportunities to automate various aspects of narrative creation \cite{kunmaran2023scenecraft}. LLMs are now able to generate coherent and contextually rich narratives, supporting the development of dynamic game storylines. Numerous studies have focused on the factors that influence the effectiveness of LLMs in narrative generation, with a focus on optimizing these models to produce high-quality stories.

For example, Harmon and Rutman's study highlights the efficacy of zero-shot prompts for character preferences and consequence prediction in improving narrative content generation \cite{harmon2023prompt}. Although their results indicate that more studies are needed for few-shot prompts, they suggest that zero-shot prompts provide successful responses in this regard \cite{harmon2023prompt}. In addition, Yuan et al. \cite{yuan2022wordcraft} presented Wordcraft, an online tool that aims to support writers by using LLMs for the creation of stories. By enhancing the collaborative experience for writers, Wordcraft demonstrated the potential of LLMs in the field of creative writing.

In a similar context, Swanson et al. \cite{swanson2021story} developed Story Centaur to provide a user interface for storywriters with limited knowledge of LLMs and prompts to interact with LLMs. Story Centaur aims to provide storywriters with a variety of tips on how to create a story and to support the story creation process. Lastly, a study by Taveekitworachai et al. \cite{taveekitworachai2023breaking} focused on ChatGPT's ability to generate stories with different endings (positive and negative). Their results show that ChatGPT has a bias towards generating stories with positive endings, which it considers less harmful than stories with negative endings.

\subsection{Visual Novel Games and LLMs}\label{sec:related_work:visual_novel}

Visual novel games, distinct from other genres, are characterized by the player's decisions directly influencing the narrative, characters, and storyline progression. This genre emphasizes a narrative-centered structure, aiming to enhance the player's engagement with the story \cite{andrew2019analyzing}. Considering the importance of narrative in this game genre, recent innovations in LLMs have the potential to significantly impact story creation. Consequently, some studies in the literature have focused on the implementation of LLMs in the context of visual novel games.

Xu et al. \cite{xu2024mozart} used a visual novel game to present their multimodal music generation framework ``Mozart's Touch'', which uses LLMs. By converting a visual description from the game into a musical composition, they demonstrated the framework's ability to translate visual content into appropriate music, enhancing the emotional and atmospheric experience. Gursesli et al. \cite{gursesli2023chronicles} conducted a study using LLMs to generate two different visual novel stories about global warming. One story was created using a set of provided keywords, while the other was created without such guidance. The stories were evaluated on several criteria, including coherence, inspiration, readability, word complexity and narrative fluency. The results showed that while each story had unique strengths, the keyword-inspired story had superior coherence. In this study, unlike the research of Gursesli et al., \cite{gursesli2023chronicles} we tried to obtain better results by using our novel strategy DCP/P.

\section{Framework}\label{sec:framework}

In this section, we discuss our framework for generating a visual novel game using generative AIs. We detail our proposed approach to prompting/programming LLMs, DCP/P, in \Cref{sec:framework:dcpp}. Additionally, details on our approach to handling context overflow and our baseline approach can be found in \Cref{sec:framework:dcpp:context} and \Cref{sec:framework:dcpp:baseline}, respectively. We provide additional details on prompts, supporting systems of the algorithm, and the web game interface in the \hyperref[appendix]{Appendix}.

\subsection{DCP/P Approach}\label{sec:framework:dcpp}

\begin{figure*}
    \centering
    \includegraphics[width=\linewidth]{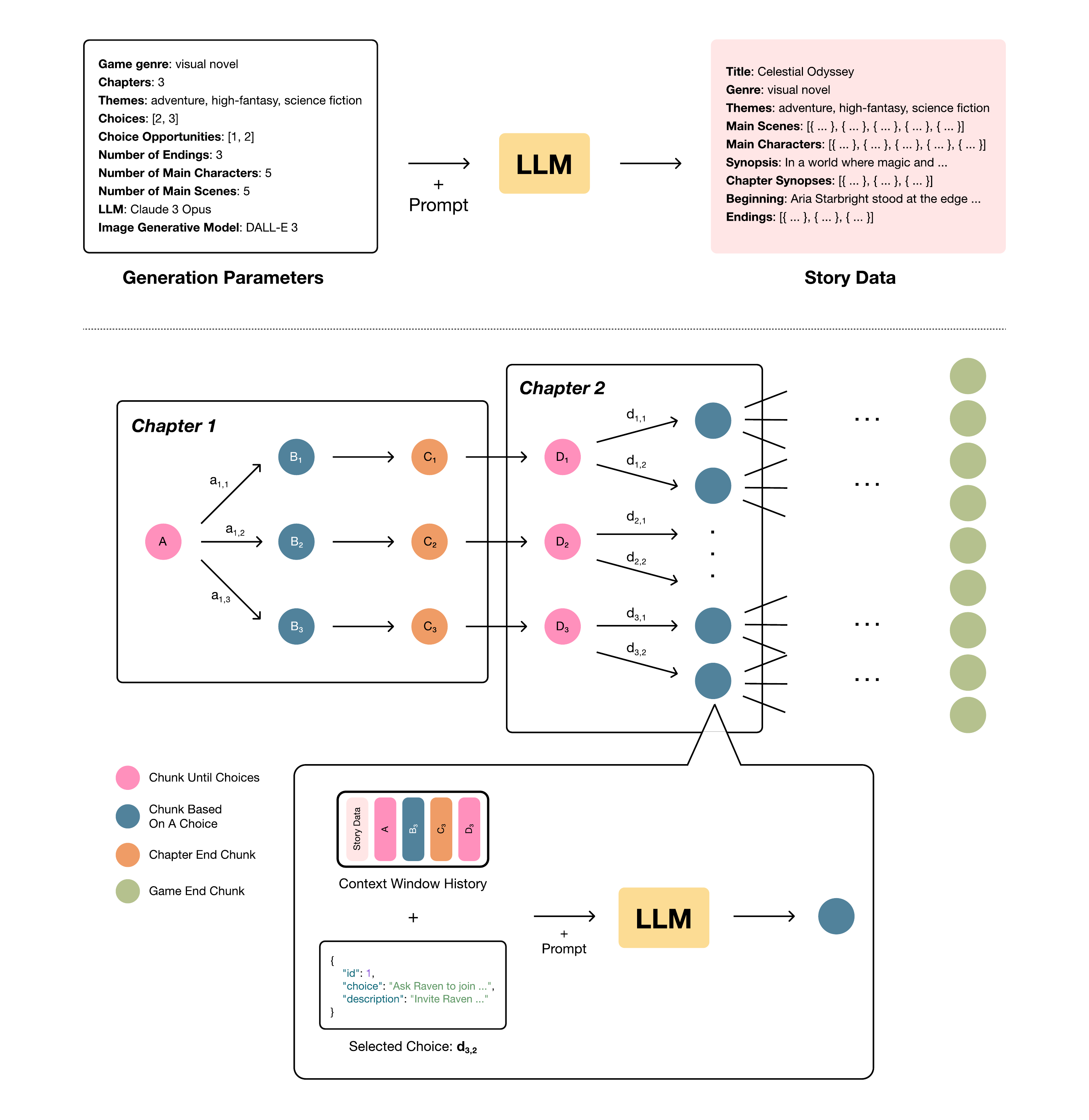}
    \caption{An overview of our framework showing the generation of story data (top) and a story chunk using the DCP/P approach (bottom).}
    \label{fig:overview}
\end{figure*}

An overview of our approach is shown in \autoref{fig:overview}. The first step of the algorithm is to have an LLM generate story data consisting of seven essential narrative elements: the story title, main scenes (locations), main characters, a story synopsis, chapter synopses, story beginning, and story endings. We allow the LLM to independently generate each element with minimal interference, guiding the generation only with themes, game genre, and other parameters determining the size of the story. These parameters include the minimum and maximum numbers of choices and choice opportunities, number of chapters, number of endings, number of main characters, and number of main scenes. The prompt used to generate the story data, along with the generated story data, serves as the initial messages in the context window history of the LLM. Once the story data are successfully generated, the process proceeds to generating story chunks.

Our DCP/P approach draws inspiration from dynamic programming and breadth-first search methodologies. Each iteration involves generating a story chunk through LLM interaction with the conversation history of the parent chunk, which varies based on each story branch. This dynamic context history is inspired by dynamic programming, where the algorithm continuously expands the solution frontier by solving subproblems. Similarly, our DCP/P approach continues to cover subparts of the story by generating story chunks, treating each chunk as a subproblem. A prompt is dynamically chosen based on the current branching type attached to the chunk-to-be-generated information in a frontier.

The branching types consist of normal, end-of-chapter, and end-of-game branching. Each story chunk encapsulates crucial narrative elements: the current story progression, detailed narrative events, and decision points that shape the narrative direction. After the story chunk is generated, a list of children is inserted into the processing queue, determined by its branching type. The conversation history used in generating this story chunk is saved within the story chunk data for generating its subsequent children chunks.

\begin{algorithm*}[htbp]
\caption{DCP/P}
\SetAlgoLined
\KwData{$generation\_config$}
\KwResult{a visual novel game based on the $generation\_config$ from $chunks$ and associated $relationships$}

$initial\_chunk \gets initialize\_generation(generation\_config)$\;
$frontiers \gets [initial\_chunk]$\;

\While{$frontiers\ \textbf{is}\ \textbf{not}\ empty$}{
    $current\_chunk \gets frontiers.pop()$\;
    $num\_choices \gets random\_int(min\_choice\_opps,\ max\_choice\_opps)$\;

    \Switch{$current\_chunk.status$}{
        \Case{$BRANCHING\_WITHOUT\_CHOICE$}{
            $new\_chunk \gets generate\_chunk\_until\_choice\_opp(current\_chunk)$\;
        }
        \Case{$BRANCHING\_WITH\_CHOICE$}{
            $new\_chunk \gets generate\_chunk\_until\_choice\_opp(current\_chunk,$\\ \ \ $current\_chunk.selected\_choice)$\;
        }
        \Case{$CHAPTER\_END$}{
            $new\_chunk \gets generate\_chunk\_until\_chapter\_end(current\_chunk)$\;
        }
        \Case{$GAME\_END$}{
            \textbf{continue}\;
        }
    }
    $frontiers.push(new\_chunk)$\;
}
\end{algorithm*}

\subsubsection{Context Overflow Policy}\label{sec:framework:dcpp:context}

The context overflow policy in our framework plays a crucial role in managing conversation histories efficiently while adhering to token limits. LLMs' token limits are determined from the pre-training time and are an inherent limitation. Therefore, this policy employs a rolling history mechanism to monitor and adjust the accumulated token count within the context of ongoing interactions. Initially, the policy calculates the total token count of the conversation history. If this count exceeds 80\% of the maximum allowable tokens, indicating potential overflow, the policy initiates a selective history truncation process.

During truncation, the policy prioritizes retaining recent user messages, ensuring essential context is preserved. It starts from the latest user message and iteratively checks the format and order of preceding assistant messages to maintain conversation coherence. The policy also includes a few initial chunks of story data to provide enough contextual information about the game.

Furthermore, to optimize resource utilization and ensure sustained performance, the policy limits the total token count to 60\% of the maximum tokens after truncation. This threshold ensures that sufficient context is retained. The retained history is integrated back into the conversation flow, and the normal generation process continues.

\subsubsection{Baseline Approach}\label{sec:framework:dcpp:baseline}

In contrast to our DCP/P approach, the baseline method simplifies the generation process by utilizing only the initial history from the story data generation phase for each interaction with the LLM. Unlike our method, which saves the conversation history within each story chunk generation, the baseline approach does not maintain a detailed record of previous interactions.

\section{Objective Evaluation}\label{sec:obj_eval}

We follow an existing study \cite{gursesli2023chronicles} for objective evaluation of five linguistic aspects of generated stories: 1) coherence, 2) inspiration, 3) narrative fluency, 4) readability, and 5) word complexity. The previous study proposed utilizing LLMs as judges to evaluate narrative content, using generated knowledge prompting \cite{liu2022generated} to condition the LLM to evaluate based on specific criteria for each aspect. We adopt the same evaluation settings and reuse their prompts for evaluating our stories.

However, our game is more complex than the one in the previous study, as each game consists of multiple story chunks across all story branches. Therefore, we evaluate each story chunk first and then average the objective scores for each aspect. Finally, we compute an average score for each aspect to represent the final standing of each game. We evaluate each game only once, as the sampling temperature is already set to 0 for deterministic behavior, as in the previous study.

\section{Experimental Setup}\label{sec:exp}

We generated a total of 20 stories, with 10 stories for each approach. Each story was generated using three specified themes: ``adventure'', ``high-fantasy'', and ``science fiction''. We selected these themes because they are broad enough for LLMs to be creative about the story, and these LLMs should have a sufficient understanding of these concepts from their training data. The remaining parameters for generating our visual novel games are as follows: three chapters, three endings, five main characters, and five main scenes. We arbitrarily selected these values to ensure an engaging and adequately lengthy story, balancing depth and complexity without overwhelming the generative model.

To reduce generation costs, we limited the number of choice opportunities for each chapter to a minimum of 1 and a maximum of 2. To mitigate biases from the story data, we also cross-initialized the story data. This means that there are a total of 10 unique story data sets, and we generated two games from each story data set: one using the baseline approach and another using the DCP/P approach. We utilized Claude 3 Opus as the LLM to generate and decide on each story transition point, choices, casts, scenes, and endings.

\section{Results and Discussions}\label{sec:result_dis}

\begin{figure}[htbp]
  \centering
  \includegraphics[width=\linewidth]{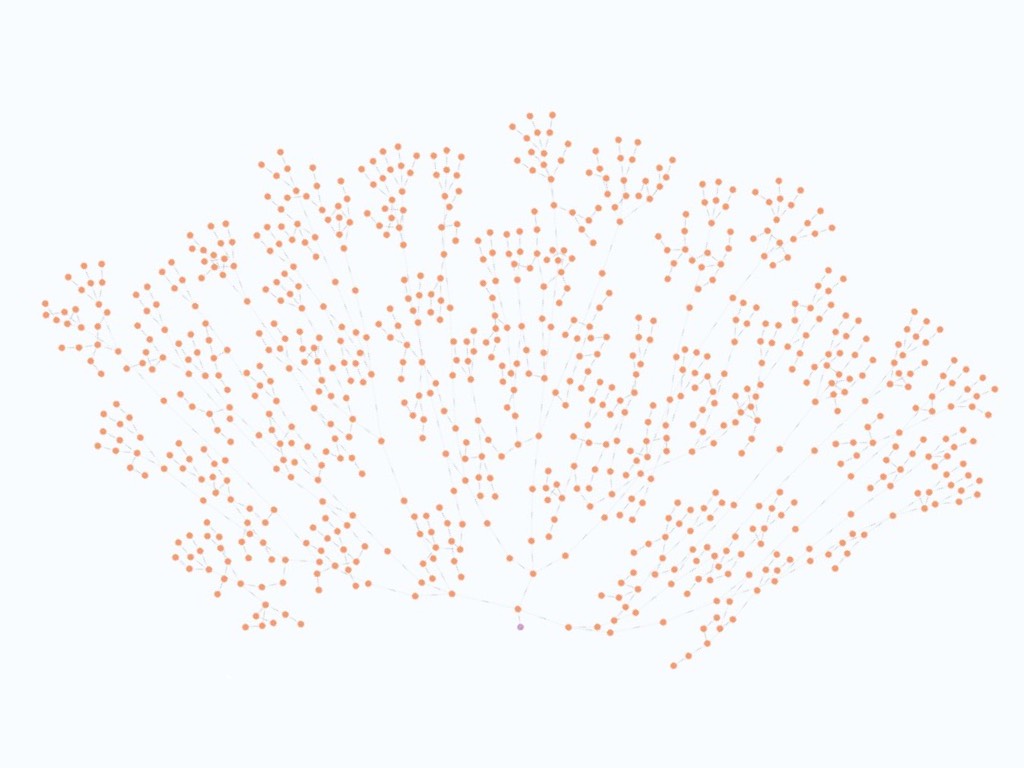}
  \caption{These figures show all nodes representing each chunk of the game story of the best-performing game generated using DCP/P. The purple node is the root node representing the story data, and the immediately connected node is the first chunk, i.e., the beginning of the story. All leaf nodes represent possible endings of the story.}
  \label{fig:nodes}
\end{figure}

We generated a total of 20 games using the method described in \Cref{sec:exp}. Each game is stored in a graph database and can be visualized similar to what is shown in \autoref{fig:nodes}. Across all approaches, there are 649.55 $\pm$ 36.49 story chunks per game on average. Separately, there are 577.7 $\pm$ 53.29 story chunks on average for games generated using the baseline approach and 721.4 $\pm$ 40.45 story chunks for games generated using the DCP/P approach.

It is evident that games generated using the DCP/P approach have a higher number of story chunks on average. This difference is due to the fact that the DCP/P approach enables LLMs to better follow instructions in the prompts and reduces the likelihood of deviation from the instructions. We observed instances where the model followed our instruction to generate a specific number of choices—for example, instructing it to generate three choices but only generating one. This occurred more frequently when using the baseline approach. Therefore, the number of choices at each choice opportunity, i.e., the branching factor, is reduced, resulting in a decreased number of chunks.

This behavior occurs because the generation model lacks enough context to fully understand the current state of the game and what to generate next, causing it to miss its generation targets. Furthermore, instructing the model to generate structured output in JSON also contributes to this issue, as the model needs to decide and generate not only the content but also syntactical symbols. Contrary to our expectation that generating content using a longer context may confuse the LLM, it actually helps the LLM follow our instructions better.

There are two main reasons why the DCP/P approach is better than the baseline. First, LLMs are conditional probabilistic \cite{bengio2000neural} autoregressive \cite{touvron2023llama} generative models. The generation of each output token is conditioned by the given input token, i.e., a prompt, and the generated tokens so far. Naturally, with more relevant tokens available to condition the LLMs, they can generate more relevant and related content. Hence, enabling LLMs to focus their attention on following instructions better with localized output probability distribution. Compared to the baseline approach that lacks context, LLMs are more prone to a probability distribution that is less localized and with bias from JSON-related tokens for formatting, there is a higher chance that the model will stop early, i.e., generating the `\}' token and stopping token too early (not yet finishing generating all choices).

While giving a longer context provides a stronger conditional strength to LLMs, LLMs are known to be prone to the symptom called \textit{lost in the middle} in long context scenarios \cite{liu2023lost}, and training LLMs to support long context is non-trivial due to various constraints \cite{chen2023extending}. However, with advancements in the field in recent years, there are studies \cite{an2024make,chen2023extending,ding2024longrope} aiming to mitigate such issues. Claude 3 Opus, which we utilize as a generation model for this study, is known to be one of the models that support long context and excel in maintaining its ability to understand such long context \cite{anthropic2024claude}. This innate ability of Claude 3 Opus is another important factor contributing to the success.

\subsection{Objective Evaluation}\label{sec:result_dis:obj_eval}

We performed objective evaluation as described in \Cref{sec:obj_eval}. Out of 20 games, two games were unsuccessful in the objective evaluation, one from each approach. This was due to improperly formatted responses causing the evaluation model to fail to continue. Therefore, we present the objective evaluation results of the remaining 18 games, nine from each approach, in \autoref{tab:obj_results}.

\begin{table*}[htbp]
    \setlength{\tabcolsep}{4pt}
    \renewcommand{\arraystretch}{1.3}
    \centering
        \caption{This table presents the objective results of 18 games, nine from each approach. \textbf{ID} denotes the first four characters of a unique ID generated for each game. \textbf{SD} denotes the story data used to generate each game, where the same character denotes the same story data. \textbf{Approach} denotes the approach used for generating a game. \textbf{Avg. Score} is the average score and its standard deviation across all linguistic aspects from all story chunks of the same game. \textbf{Cohe.}, \textbf{Insp.}, \textbf{Narr.}, \textbf{Read.}, and \textbf{Word.} denote the average score and associated standard deviation across all story chunks of the same game for coherence, inspiration, narrative fluency, readability, and word complexity, respectively.}
        \label{tab:obj_results}
        \begin{tabularx}{\linewidth}{@{}r*9{>{\centering\arraybackslash}X}@{}}
        \hline
        \textbf{ID} & \textbf{Approach} & \textbf{SD} & \textbf{Avg. Score} & \textbf{Cohe.} & \textbf{Insp.} & \textbf{Narr.} & \textbf{Read.} & \textbf{Word.}\\
        \hline
        fff6 & DCP/P & A & \textbf{6.81} $\pm$ 0.36 & \textbf{7.20} $\pm$ 0.61 & \textbf{6.34} $\pm$ 0.56 & \textbf{6.86} $\pm$ 1 & \textbf{7.22} $\pm$ 0.51 & 6.46 $\pm$ 0.58 \\
        61b6 & DCP/P & D & 6.76 $\pm$ 0.31 & 7.07 $\pm$ 0.58 & 6.24 $\pm$ 0.44 & 6.82 $\pm$ 0.85 & 7.06 $\pm$ 0.45 & \textbf{6.62} $\pm$ 0.54 \\
        6a8b & DCP/P & I & 6.72 $\pm$ 0.34 & 7.12 $\pm$ 0.56 & 6.25 $\pm$ 0.61 & 6.66 $\pm$ 0.97 & 7.09 $\pm$ 0.55 & 6.48 $\pm$ 0.59 \\
        4aa9 & DCP/P & F & 6.69 $\pm$ 0.39 & 7.10 $\pm$ 0.59 & 6.14 $\pm$ 0.55 & 6.74 $\pm$ 0.86 & 7.11 $\pm$ 0.48 & 6.37 $\pm$ 0.53 \\
        c185 & DCP/P & C & 6.67 $\pm$ 0.31 & 6.92 $\pm$ 0.63 & 6.22 $\pm$ 0.44 & 6.62 $\pm$ 0.91 & 7.09 $\pm$ 0.56 & 6.5 $\pm$ 0.55 \\
        ea72 & DCP/P & H & 6.61 $\pm$ 0.31 & 6.87 $\pm$ 0.68 & 6.17 $\pm$ 0.60 & 6.54 $\pm$ 0.92 & 7.05 $\pm$ 0.46 & 6.43 $\pm$ 0.59 \\
        3bd1 & DCP/P & J & 6.61 $\pm$ 0.34 & 7 $\pm$ 0.61 & 6.16 $\pm$ 0.41 & 6.49 $\pm$ 1.03 & 6.99 $\pm$ 0.54 & 6.38 $\pm$ 0.5 \\
        3859 & DCP/P & B & 6.58 $\pm$ 0.35 & 6.79 $\pm$ 0.84 & 6.14 $\pm$ 0.6 & 6.52 $\pm$ 0.94 & 7.12 $\pm$ 0.5 & 6.31 $\pm$ 0.55 \\
        790a & DCP/P & E & 6.56 $\pm$ 0.31 & 6.79 $\pm$ 0.73 & 6.16 $\pm$ 0.52 & 6.55 $\pm$ 0.91 & 7.01 $\pm$ 0.63 & 6.3 $\pm$ 0.49 \\
        d979 & Baseline & D & 6.53 $\pm$ 0.33 & 6.83 $\pm$ 0.61 & 6.02 $\pm$ 0.53 & 6.49 $\pm$ 0.94 & 6.96 $\pm$ 0.57 & 6.37 $\pm$ 0.51 \\
        8f89 & Baseline & G & 6.4 $\pm$ 0.41 & 6.76 $\pm$ 0.62 & 5.91 $\pm$ 0.55 & 6.25 $\pm$ 0.91 & 7 $\pm$ 0.57 & 6.10 $\pm$ 0.39 \\
        0f83 & Baseline & E & 6.39 $\pm$ 0.34 & 6.61 $\pm$ 0.64 & 6.04 $\pm$ 0.48 & 6.21 $\pm$ 0.92 & 6.95 $\pm$ 0.63 & 6.15 $\pm$ 0.44 \\
        5f2d & Baseline & F & 6.35 $\pm$ 0.41 & 6.55 $\pm$ 0.73 & 5.80 $\pm$ 0.63 & 6.26 $\pm$ 0.93 & 7.01 $\pm$ 0.56 & 6.14 $\pm$ 0.46 \\
        c5b4 & Baseline & H & 6.30 $\pm$ 0.3 & 6.44 $\pm$ 0.81 & 5.86 $\pm$ 0.62 & 6.22 $\pm$ 0.9 & 6.78 $\pm$ 0.7 & 6.21 $\pm$ 0.5 \\
        681f & Baseline & B & 6.27 $\pm$ 0.39 & 6.42 $\pm$ 0.79 & 5.67 $\pm$ 0.71 & 6.23 $\pm$ 0.91 & 6.87 $\pm$ 0.71 & 6.19 $\pm$ 0.5 \\
        c249 & Baseline & A & 6.27 $\pm$ 0.38 & 6.4 $\pm$ 0.82 & 5.82 $\pm$ 0.61 & 6.09 $\pm$ 0.93 & 6.92 $\pm$ 0.66 & 6.09 $\pm$ 0.46 \\
        0a80 & Baseline & C & 6.26 $\pm$ 0.29 & 6.34 $\pm$ 0.86 & 5.89 $\pm$ 0.49 & 6.09 $\pm$ 0.92 & 6.76 $\pm$ 0.75 & 6.25 $\pm$ 0.44 \\
        57a4 & Baseline & J & 6.25 $\pm$ 0.33 & 6.52 $\pm$ 0.79 & 5.87 $\pm$ 0.56 & 5.99 $\pm$ 0.96 & 6.74 $\pm$ 0.77 & 6.12 $\pm$ 0.45 \\
        \hline
        \end{tabularx}
\end{table*}

The game with the highest average objective score is from the DCP/P approach, scoring 6.81 $\pm$ 0.36, outperforming the best-performing game from the baseline approach, which scored 6.53. The best game from the DCP/P approach also holds the highest scores in all linguistic aspects except for word complexity. The highest word complexity score is held by the first runner-up, also from the DCP/P approach. Additionally, all games generated using the DCP/P approach perform better than the best-performing game using the baseline approach, considering the average objective scores. This observation clearly demonstrates the superiority of the DCP/P approach.

We also observe that, in general, games generated from the DCP/P approach have high coherence scores. This is anticipated as DCP/P aims to provide the generative model with previous context to generate more coherent story chunks. We discuss this aspect further in the next section on qualitative quality. This trend extends to the inspiration aspect as well. For the inspiration aspect, we aim to assess how consistently the generated game aligns with the specified themes provided at the beginning of the context during the generation process. We observe that DCP/P is better at drawing inspiration from the themes compared to the baseline approach.

Narrative fluency and readability are related aspects, and we also see that DCP/P-generated games generally have higher scores in these areas as well. The results show that DCP/P not only effectively reminds the model of previous content but also aids the LLM in generating more fluent narrative elements. For word complexity, we observe generally equivalent results from both approaches, with games from the DCP/P approach having slightly higher scores. Therefore, we conclude that word complexity is more related to the LLMs' innate preferences and less dependent on whether it knows the previous context or not.

\subsection{Qualitative Analysis}\label{sec:result_dis:qual}

While objective evaluation is useful in providing a numerical metric to understand the generated content, it falls short in helping us understand the actual qualitative quality of the generated game. Therefore, we conducted a qualitative analysis by closely examining a game with the top score from each approach. We also provide a title and synopsis generated by the model for each game to give an overview of the story. Then, we discuss each game from each approach in subsequent sections.

Through our testing, we found that each game generally takes about 15 minutes to play through. The image assets, characters, and scenes are quite impressive. However, the generated images are not perfect. There are instances where the generated characters contain decorative elements, resulting in the background removal model failing to remove these distracting elements. Cases of malformed shapes appearing in the generated scenes or artifacts and solid color borders, such as in \autoref{fig:screenshot_baseline}, are also not uncommon.

\subsubsection{Baseline: The Chronicles of Zephyr}\label{sec:result_dis:qual:baseline}\hfill\\

The best-performing game generated by the baseline approach, with an ID starting with \texttt{d979}, has the title ``\textbf{The Chronicles of Zephyr}.'' We show screenshots of this game in \autoref{fig:screenshot_baseline}. A synopsis of this game is provided in \Cref{appendix:synopsis}.

\begin{figure}[btp]
    \centering
    \begin{subfigure}[tbp]{\linewidth}
         \centering
         \includegraphics[width=\textwidth]{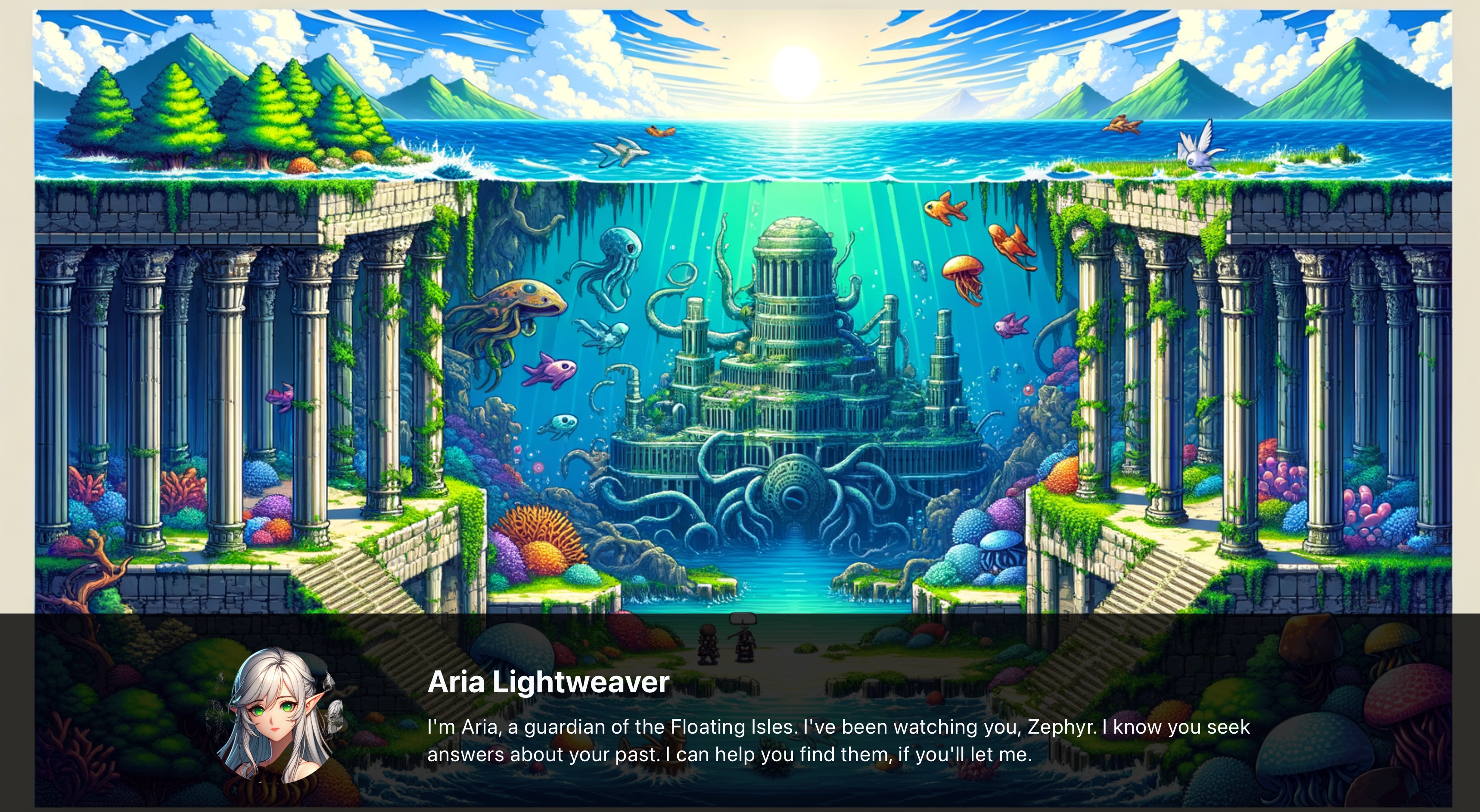}
     \end{subfigure}
    \begin{subfigure}[tbp]{\linewidth}
         \centering
         \includegraphics[width=\textwidth]{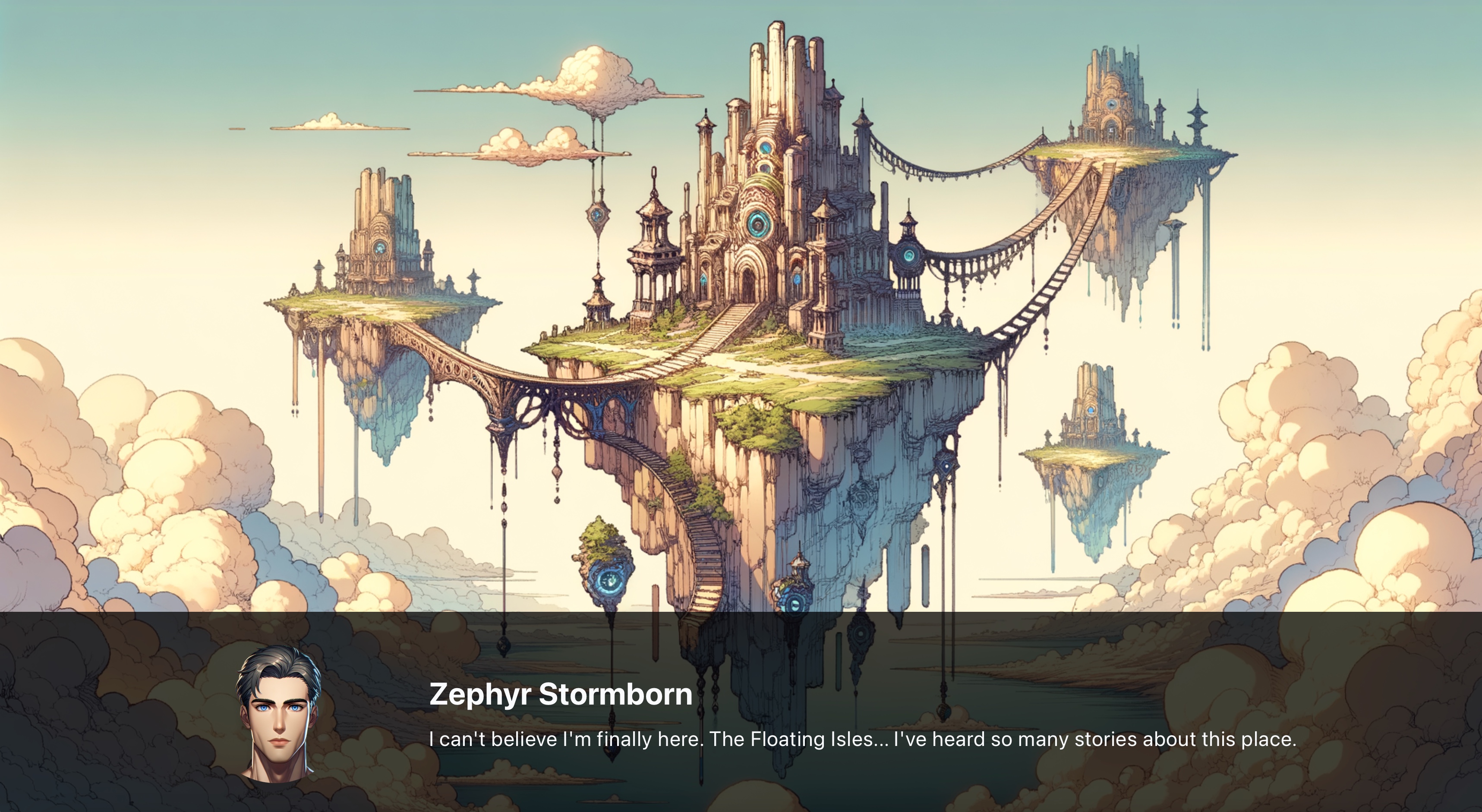}
     \end{subfigure}
     \caption{Screenshots of the best game generated using the baseline approach. Each screenshot is taken from various parts of the story during its progression.}
     \label{fig:screenshot_baseline}
\end{figure}

We observe that, in general, the story progresses according to the synopsis. However, there are several issues with the coherence of the story. The narrative may progress in a disconnected manner; for example, while the protagonist and the crew are in one place, such as the Ancient Library, they may suddenly jump to another location, like a Sunken City, in the next scene without a proper connection or transition. This sudden transition results in an abrupt and disjointed experience. Moreover, various characters may appear suddenly without a proper introduction. These observations indicate that the LLM did not understand the context well enough to craft a coherent story, which is also reflected in the lower coherence score.

Furthermore, the LLM is apparently confused regarding the matching between speakers and dialogue lines. For instance, a dialogue line that is supposed to be spoken by Aria, hinted at by the mention of Zephyr's name, is instead assigned to Zephyr. This issue also occurs when referencing certain locations, but the chosen scene does not match up. This extends to mentioning characters' names that never exist in the story, such as Aurora and Kai, causing confusion and showing signs of hallucination \cite{xu2024hallucination}. It is clear that without sufficient context, the LLM is not only unable to generate a coherent story but also struggles with smaller narrative elements and simple tasks like choosing the correct scene picture.

As previously mentioned, there also appears to be an issue with the LLM stopping the generation of choices too early, resulting in fewer choices than specified. Consequently, there are instances when a choice screen is shown, but only one choice is available for selection. Therefore, the story lacks depth in its progression.

\subsubsection{DCP/P: Celestial Odyssey}\label{sec:result_dis:qual:dcpp}

\begin{figure}[btp]
    \centering
    \begin{subfigure}[tbp]{\linewidth}
         \centering
         \includegraphics[width=\textwidth]{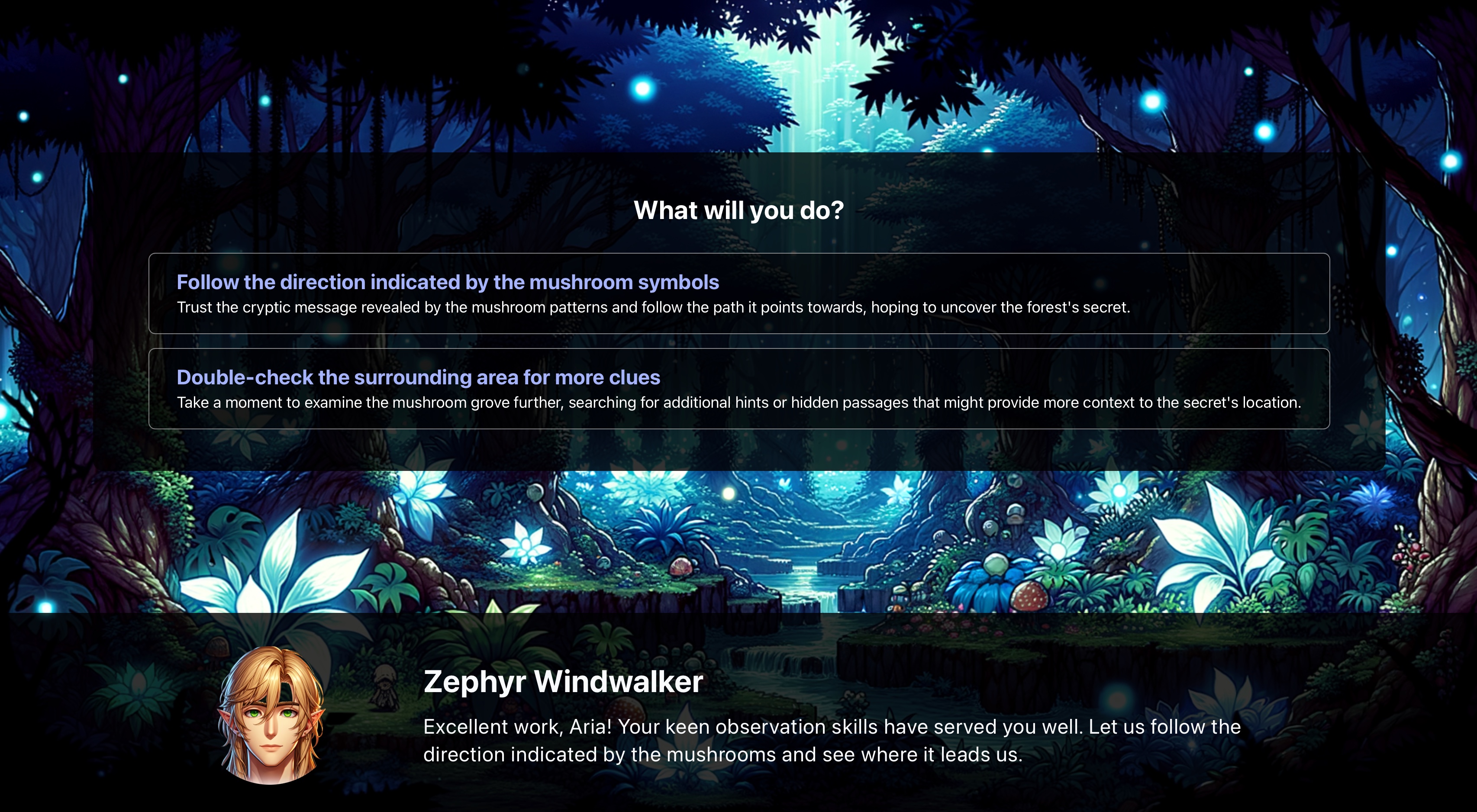}
     \end{subfigure}
    \begin{subfigure}[tbp]{\linewidth}
         \centering
         \includegraphics[width=\textwidth]{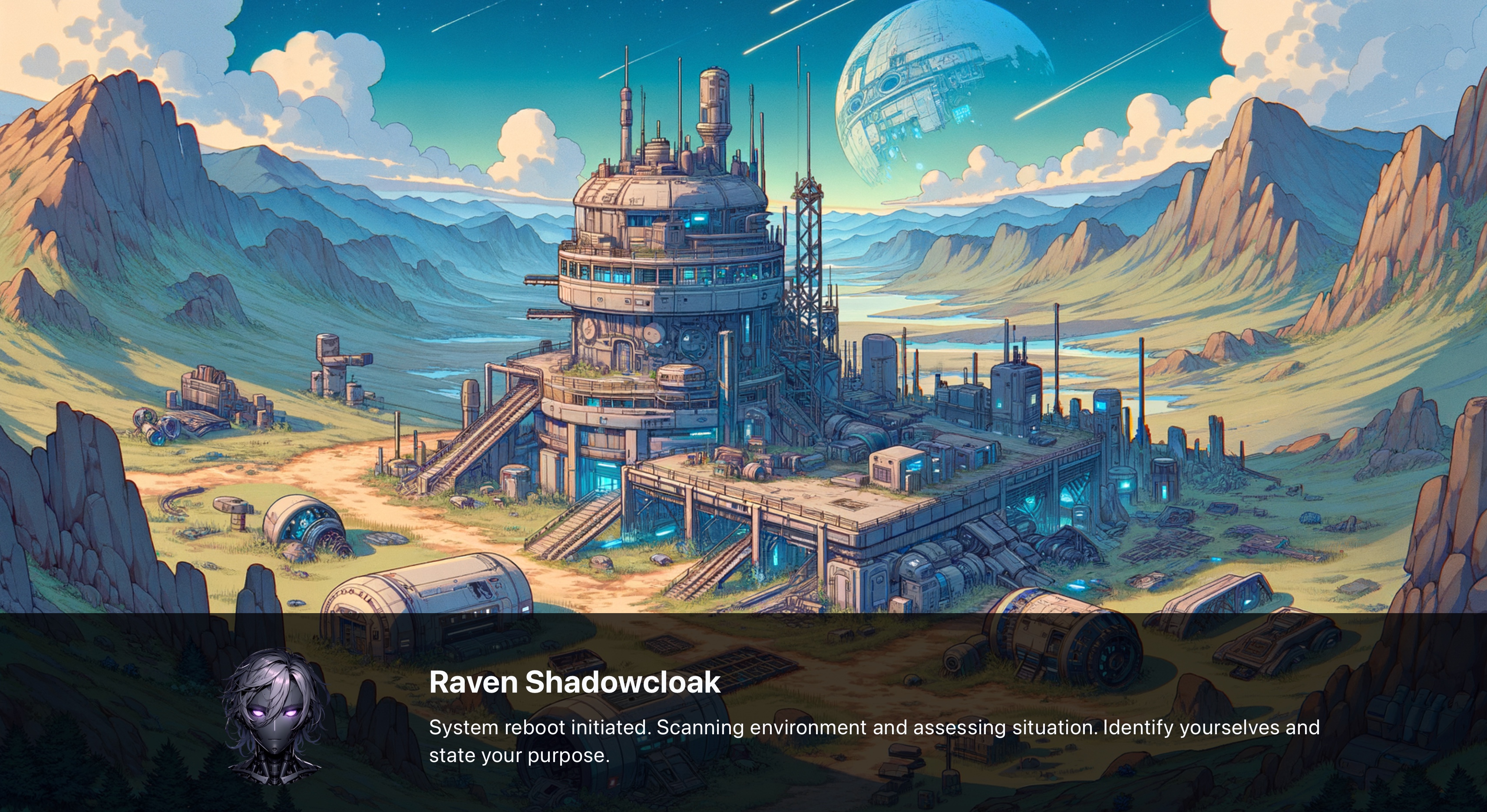}
     \end{subfigure}
     \caption{Screenshots of the best game generated using the DCP/P approach. Each screenshot is taken from various parts of the story during its progression.}
    \label{fig:screenshot_proposed}
\end{figure}

Similar to the previous subsection, we show screenshots from the best performing game generated by DCP/P approach in \autoref{fig:screenshot_proposed}. This game has an ID starts with \texttt{d979} and a title of ``\textbf{Celestial Odyssey}''. A synopsis of this game is available in \Cref{appendix:synopsis}.

First, it is worth noting that the synopsis of this game is similar to the previous one, even though they both originate from different story data. The information we conditioned to the LLM includes themes, while the rest is left open to the LLM's imagination. This similarity shows that LLMs lack creativity and are influenced by their probability distribution resulting from the training process. Similar observations were also made by previous studies \cite{taveekitworachai2023breaking,taveekitworachai2023what}. We note that we utilized a default sampling temperature for the LLM, and we may gain a more diverse response if we increase the sampling temperature. However, increasing the sampling temperature also increases the chance of generating unrelated tokens, resulting in a higher failure rate for JSON structured output generation.

Story-wise, this game has a more coherent story compared to the game generated from the baseline approach. We also observe that this game has smoother transitions and cues before moving the protagonist and the team to another location. For example, a supporting character may inform them that they need to visit a certain place to learn more about something or obtain certain artifacts. Another interesting note is that the story also progresses with an epilogue, i.e., what happens to the world after the protagonist completes their quest to save the world.

\subsubsection{Discussions}\label{sec:result_dis:qual:disc}

The story progression of the games is quite standard and involves elements commonly found in this genre. The story progresses by having the protagonist embark on a quest to restore his memory and find the true purpose of his life. Throughout the journey, they visit various places and solve problems for the citizens of these places. While there is nothing inherently wrong with this story progression style, it highlights an area where further improvement is possible.

The story style heavily depends on the prompt that instructs the LLM. In our case, we kept the initial prompt quite simple due to the nature of our work focusing on introducing the framework. To improve further, one may utilize expert prompting \cite{xu2023expertprompting} to help the LLM generate a story with more depth and an interesting narrative style. Role prompting \cite{wang2024rolellm} also shows potential in further improving the generation of more consistent dialogue and helping the LLM stay in character. We acknowledge that there is room for improvement in prompts.

Another issue that often arises in the game generated from the baseline approach is the occurrence of black screens and unknown characters. In our current framework, we instruct LLMs to only utilize existing scenes and characters that were pre-generated in the first step of the algorithm. While there are cases when LLMs exhibit their creativity by using never-before-generated characters and scenes, there are also times when they fail to select the correct character and scene, resulting in such experiences. 

One way to improve this further is to trade-off with generation cost by decomposing the task. We can first generate the dialogues and story and then have another round of LLM interaction to select the best scenes and characters or entirely generate new ones. We plan to explore this aspect further in our future work.

\subsection{Word Choice Biases}\label{sec:result_dis:choice}

\begin{figure}[tbp]
    \centering
    \begin{subfigure}[tbp]{\linewidth}
         \centering
         \includegraphics[width=\textwidth]{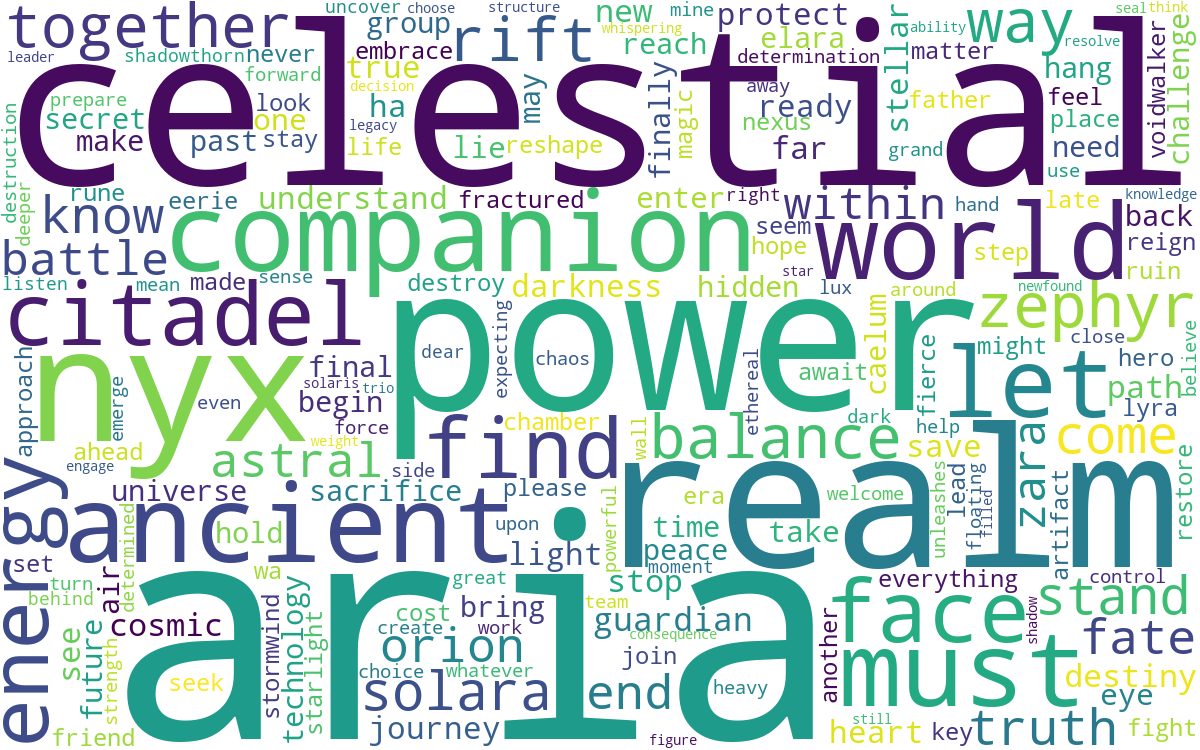}
         \caption{A word cloud showing the frequency of words from all games generated by the baseline approach.}
         \label{fig:word_cloud_baseline}
     \end{subfigure}
    \begin{subfigure}[tbp]{\linewidth}
         \centering
         \includegraphics[width=\textwidth]{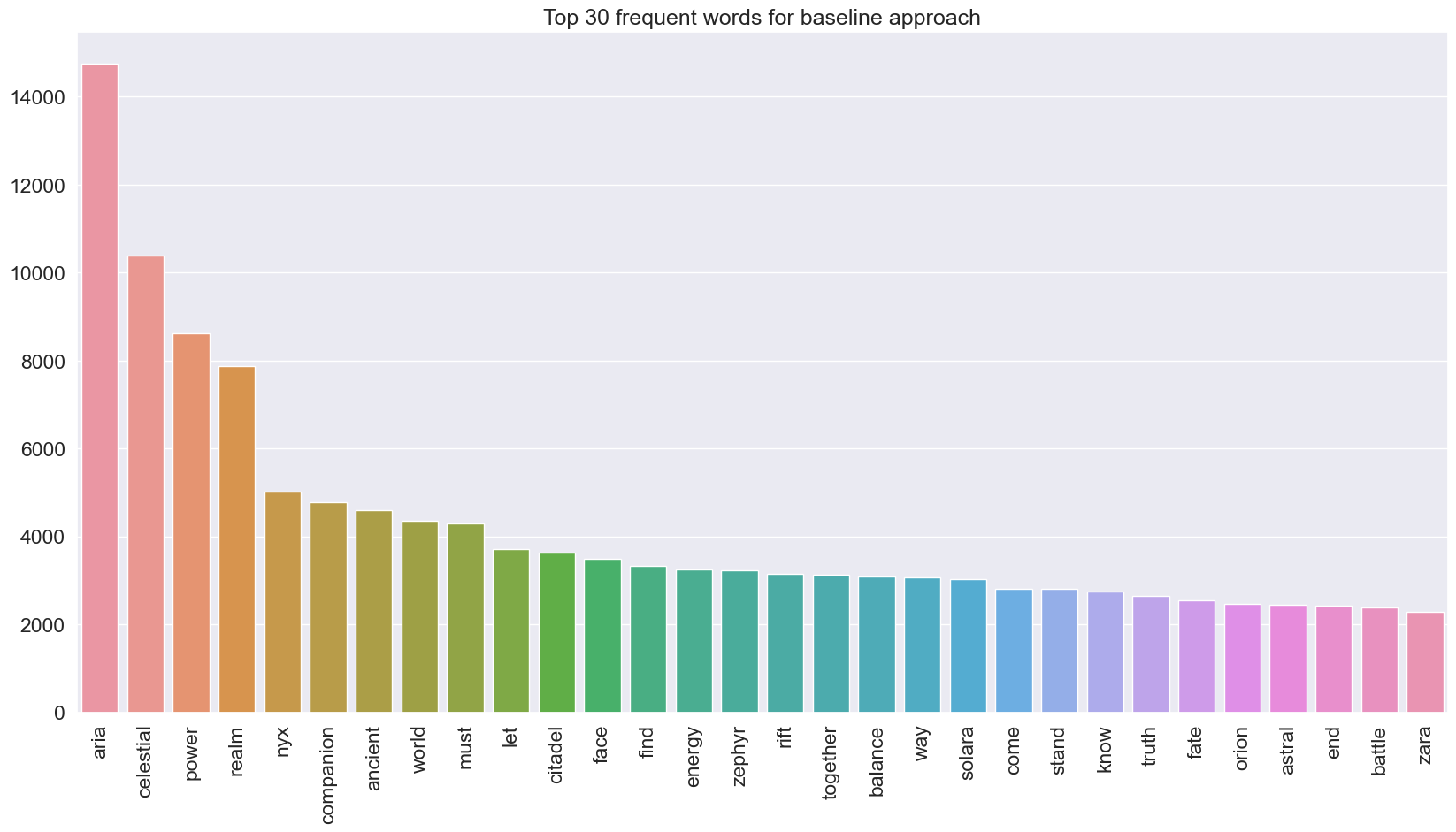}
         \caption{A bar chart showing the frequency of top 30 words from all games generated by the baseline approach.}
         \label{fig:bar_plot_baseline}
     \end{subfigure}
\end{figure}

\begin{figure}[tbp]
    \centering
    \begin{subfigure}[tbp]{\linewidth}
         \centering
         \includegraphics[width=\textwidth]{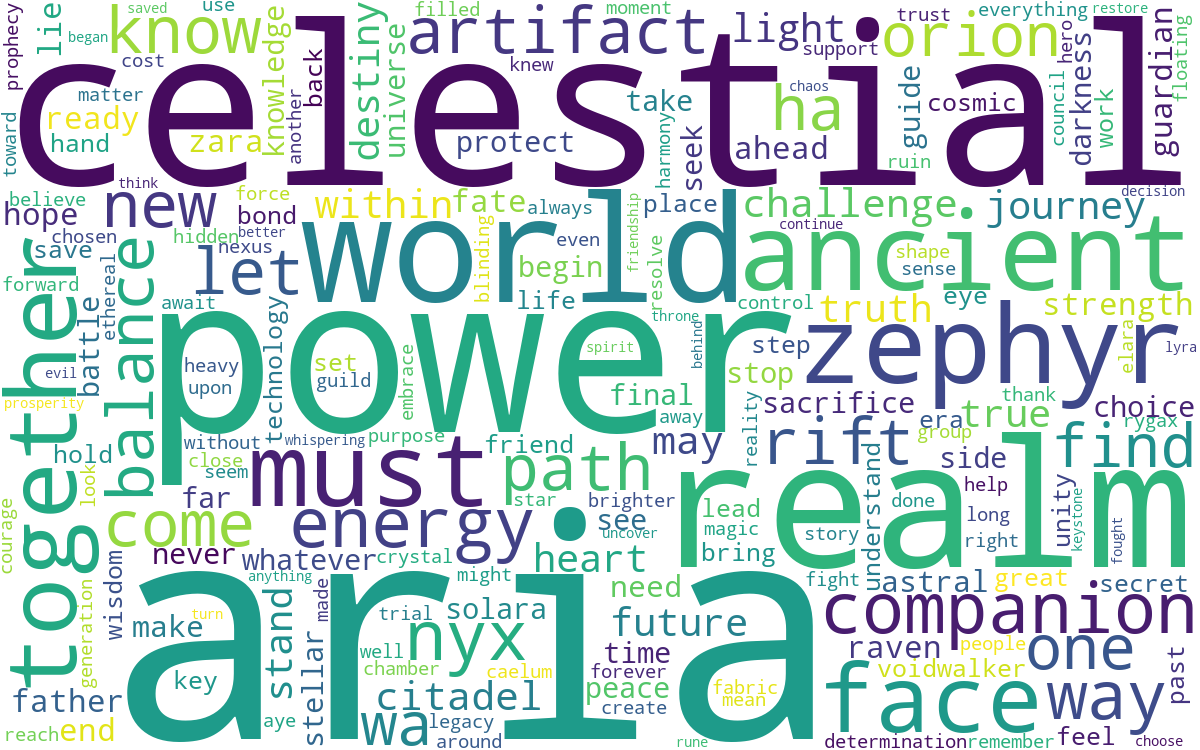}
         \caption{A word cloud showing the frequency of words from all games generated by the DCP/P approach.}
         \label{fig:word_cloud_proposed}
     \end{subfigure}
    \begin{subfigure}[tbp]{\linewidth}
         \centering
         \includegraphics[width=\textwidth]{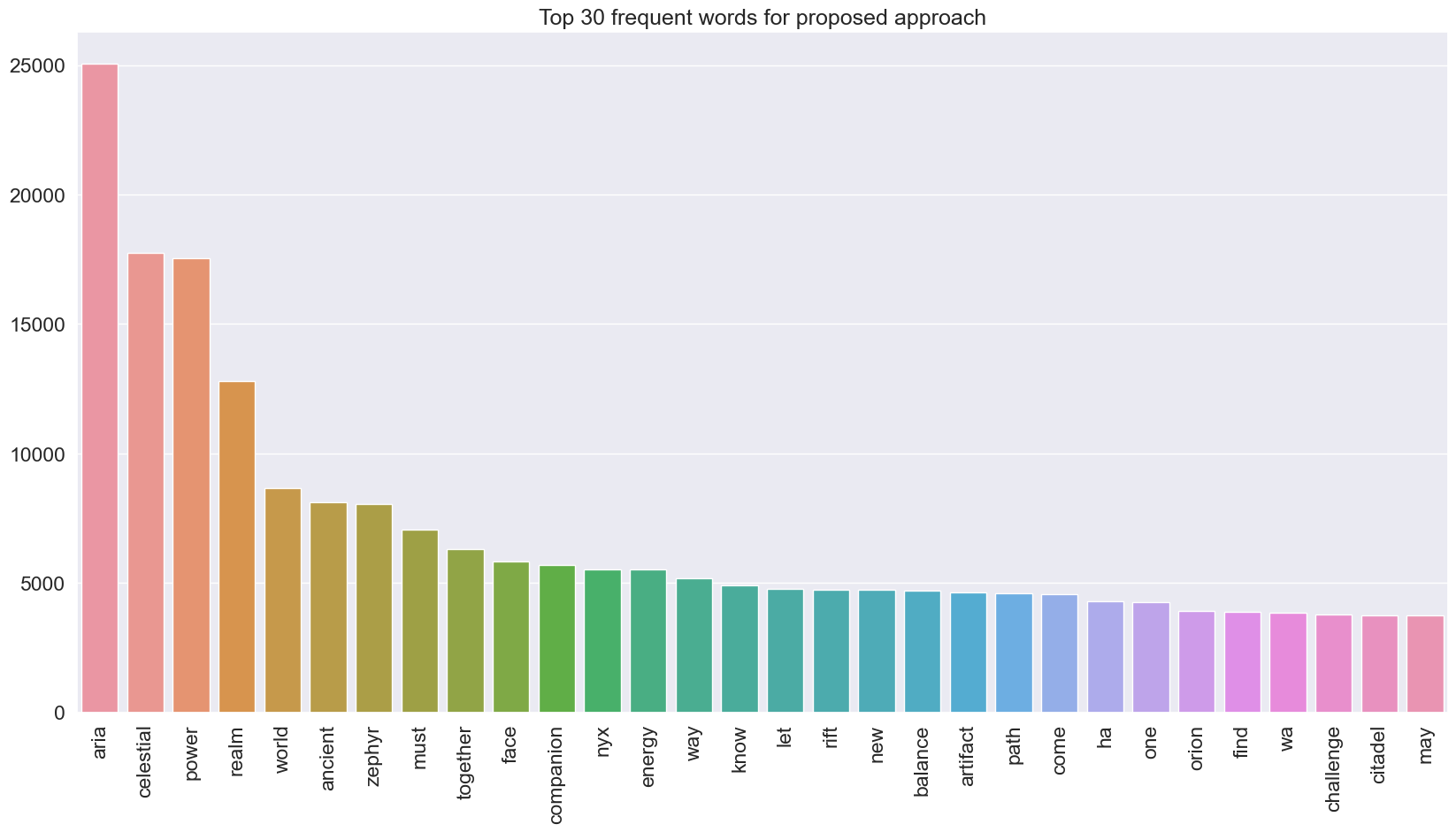}
         \caption{A bar chart showing the frequency of top 30 words from all games generated by the DCP/P approach.}
         \label{fig:bar_plot_proposed}
     \end{subfigure}
\end{figure}

We follow an existing study style of investigation \cite{taveekitworachai2023breaking,taveekitworachai2023what} on biases resulting from LLMs' training data distribution. We collect all words utilized in the games of each approach and apply standard text pre-processing steps, i.e., tokenization, and removal of single character, stop words,pych' and punctuation. We then visualize word frequency as a word cloud for each approach in Figures \ref{fig:word_cloud_baseline} and \ref{fig:word_cloud_proposed}.

Overall, we observe that word distributions of the games generated from each approach are similar. The most frequent word is ``Aria'', which should not be surprising as this character often plays an important role in the story. However, the next most frequent words show us how LLMs are biased towards very similar stories for each game. ``Celestial'' indicates that given the specified theme, the LLM always connects these to certain themes, reinforcing our previous observation that LLMs are unable to generate entirely creative and novel stories. The generated stories often reflect the most common occurrences in their training distribution.

More concerning is the appearance of the word ``Power''. We use Claude 3 Opus as our text generative model, different from previous studies \cite{taveekitworachai2023breaking,taveekitworachai2023what}, which utilized GPT-3.5 Turbo or GPT-4, but we still observe the high frequency of this word. One possible explanation is that these LLMs, even from different developers, utilize similar data sources as part of their training sets. Therefore, the issue of unoriginal content generation by LLMs cannot be easily solved by merely changing the LLM family and requires a more elaborate method.

\subsection{Word Sentiment Biases}\label{sec:result_dis:sentiment}
We also conducted an investigation on biases towards certain sentiments, i.e., positive or negative. We utilized a sentiment word list \cite{hu2004mining} that was previously used in a similar study \cite{taveekitworachai2023breaking}. We counted the number of occurrences of words from the list, either positive or negative, that appeared in the entire game across all games for each approach. We note that we pre-processed the text as described in the previous subsection before counting. We present the results in \autoref{tab:bias}.

\begin{table}[htbp]
    \setlength{\tabcolsep}{4pt}
    \renewcommand{\arraystretch}{1.3}
    \centering
        \caption{This table presents a count of positive or negative sentiment words that appeared across games for each approach.}
        \label{tab:bias}
        \begin{tabularx}{\linewidth}{@{}r*4{>{\centering\arraybackslash}X}@{}}
        \hline
        \textbf{Approach} & \textbf{\#Positive} & \textbf{\#Negative} & \textbf{\#Total}\\
        \hline
        Baseline & 35,022 (7.34\%) & 36,332 (7.62\%) & 476,827\\
        DCP/P & 79,756 (8.84\%) & 59,258 (6.57\%) & 902,095\\
        \hline
        \end{tabularx}
\end{table}

We observe that the generated games from the baseline approach consist of a similar ratio of positive and negative sentiment words, contributing to only a fraction of the total words. We observe similar ratios in games generated using the DCP/P approach. However, positive sentiment words appeared slightly more frequently than negative sentiment words in these games. Another interesting observation is that the amount of words in DCP/P games is almost twice the number of words in baseline games. This aligns with our observation that games from the DCP/P approach tend to use longer sentences, both in dialogue and description. Given that we did not change the instructions in the prompt when using the DCP/P approach, it shows an inherent bias of LLMs in generating longer content when provided with a longer context compared to not providing previous context.

We acknowledge that this analysis only provides a preliminary evaluation of sentiment biases in LLMs by counting the frequency of words. Modern techniques, such as using LLMs for sentiment analysis \cite{stigall2024large,viggiato2023leveraging}, have the potential for more accuracy and a better understanding of the context of words. However, these methods also come with a higher cost of running the analysis. Therefore, we opted for simplicity in our analysis.

\subsection{Limitations and Future Work}\label{sec:result_dis:limit}
In this paper, we demonstrate the potential of utilizing various generative models to create a visual novel game. LLMs are employed for generating text content and deciding story progression, while an image generative model is used for character and scene asset generation. We acknowledge the potential for incorporating audio generative models \cite{kreuk2023audiogen,copet2024simple} and the various aspects that could be improved in the generation process and the quality of the generated game.

We only tried Claude 3 Opus in this paper, but there are various other LLMs and image generative models available. One drawback of utilizing state-of-the-art LLMs is the long inference time, and we are intrigued by how our approach may generalize to smaller LLMs. There are reliability issues with LLMs that could be improved further, such as utilizing tools like Instructor\footnote{\url{https://github.com/jxnl/instructor}} or function calling\footnote{\url{https://platform.openai.com/docs/guides/function-calling}} for better structured output generation.

Techniques commonly deployed for LLM-integrated systems, such as retrieval-augmented generation \cite{gao2024retrievalaugmented}, are worth further exploration to personalize story branches. We also plan to conduct a subjective evaluation to understand how humans perceive the games generated by LLMs and explore different subgenres of the game. We note the potential of this algorithm to go beyond visual novel games and generate any type of graph-based content.

\section{Conclusions}\label{sec:conclusion}
We propose DCP/P, a framework for generating a visual novel game using generative models. Our framework utilizes a dynamic context to generate a coherent narrative branch at each point in the game story. Our algorithm is able to generate content and store it in a graph database, using our developed web interface to display and provide a game experience. We evaluate our proposed approach against a baseline that interacts with an LLM without providing previous context but only the story data. We show that games generated from the DCP/P approach are better than games generated using the baseline approach based on our objective evaluation of various linguistic aspects.

We also perform a qualitative analysis on the best-performing game of each approach. We observe a more coherent story and better transitions between story chunks in a game generated using the DCP/P approach. Following existing studies, we conduct word bias and sentiment analyses, finding that even with a different LLM family from previous studies, the LLM still exhibits biases towards certain words. We also provide various suggestions for further improving the DCP/P algorithm. We conclude that generative models show potential for generating a visual novel game when utilized properly.
 
\bibliographystyle{IEEEtran}
\bibliography{refs}

\section*{Appendix}\label{appendix}

\subsection{Synopsis of the Best Performing Stories}\label{appendix:synopsis}
\begin{figure}[tbp]
    \centering
     \begin{subfigure}[tbp]{\linewidth}
         \centering
         \includegraphics[width=\textwidth]{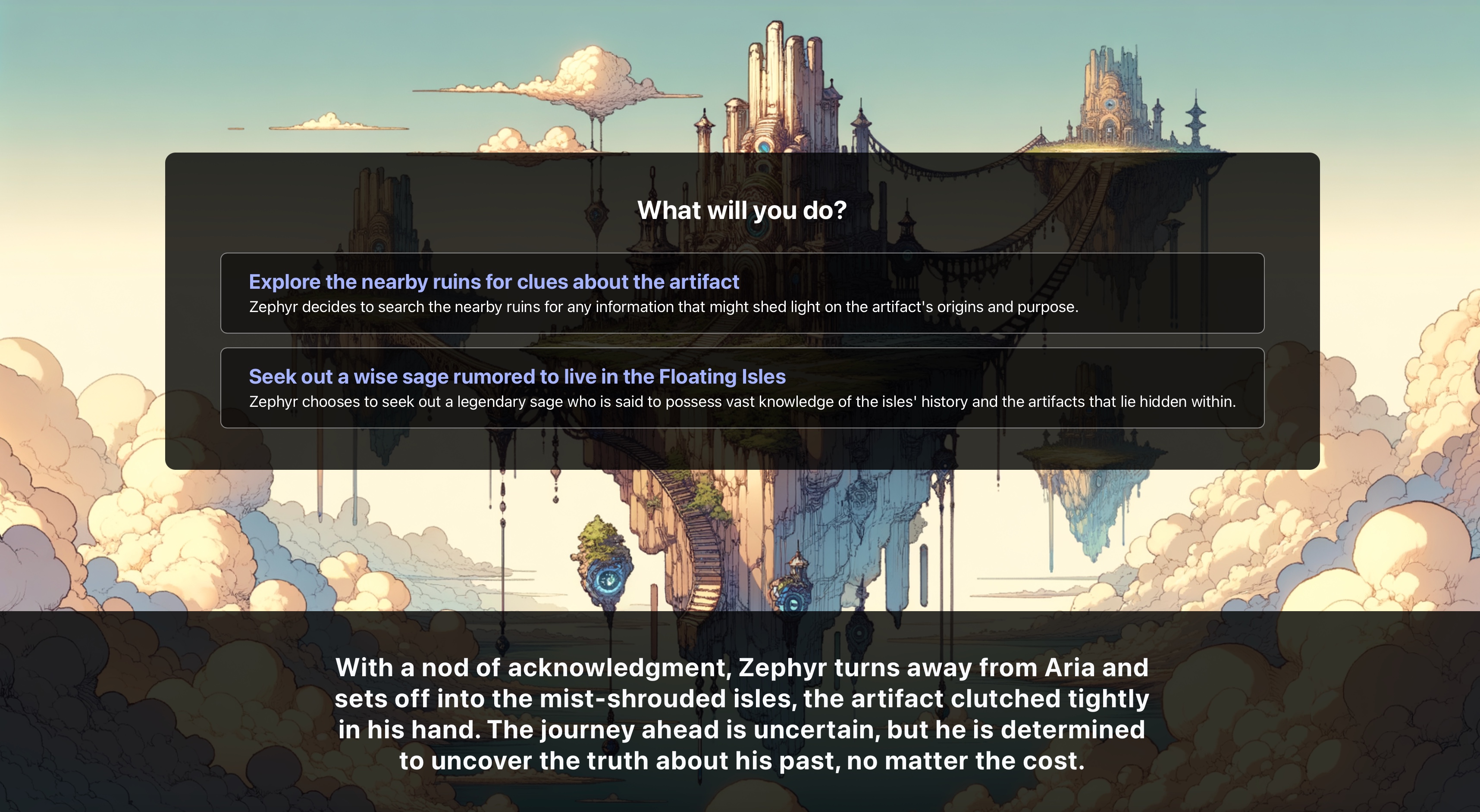}
     \end{subfigure}
    \begin{subfigure}[tbp]{\linewidth}
         \centering
         \includegraphics[width=\textwidth]{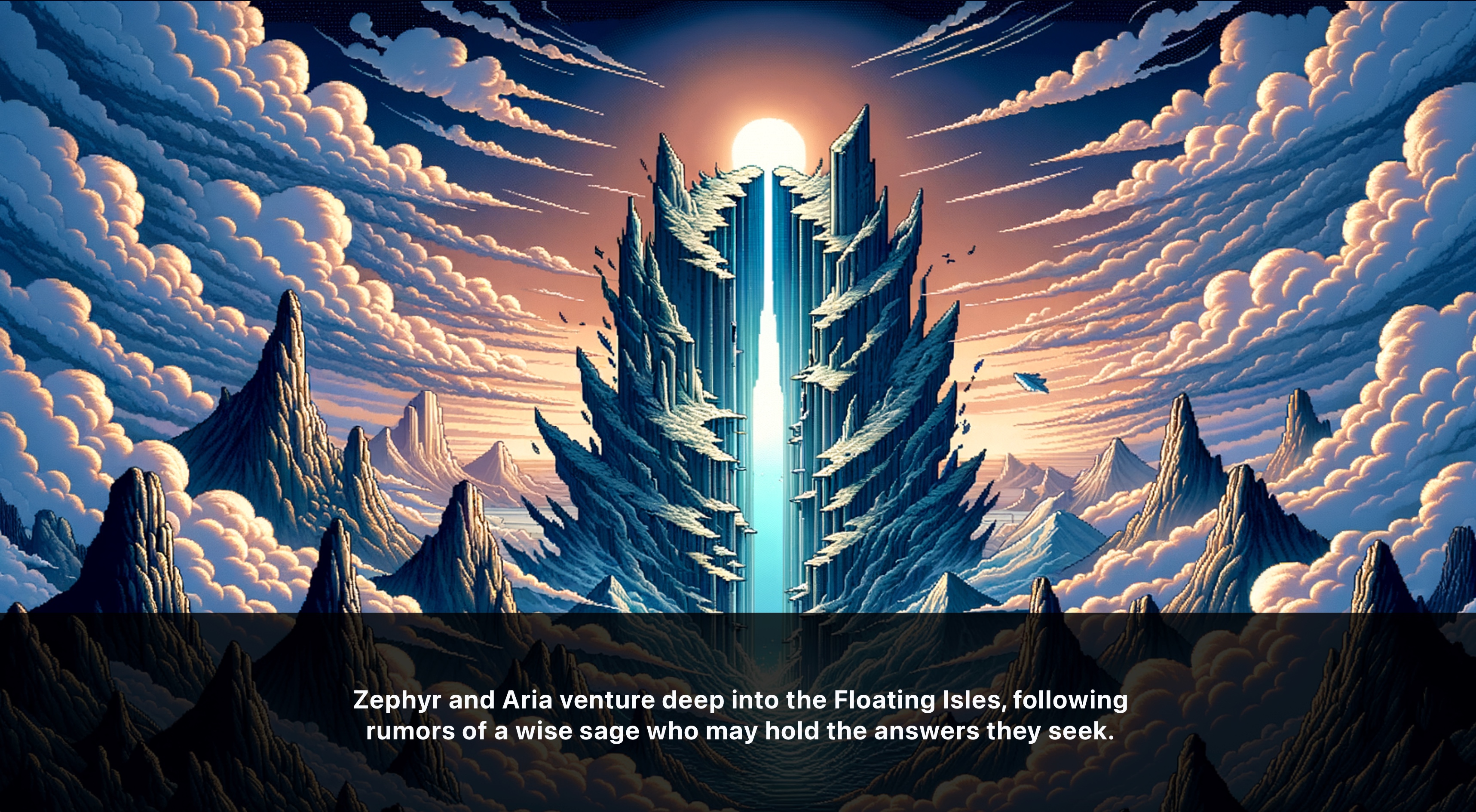}
     \end{subfigure}
     \caption{Additional screenshots of the best game generated using the baseline approach. Each screenshot is taken from various parts of the story during its progression.}
     \label{fig:additional_screenshot_baseline}
\end{figure}

\begin{figure}[tbp]
    \centering
     \begin{subfigure}[tbp]{\linewidth}
         \centering
         \includegraphics[width=\textwidth]{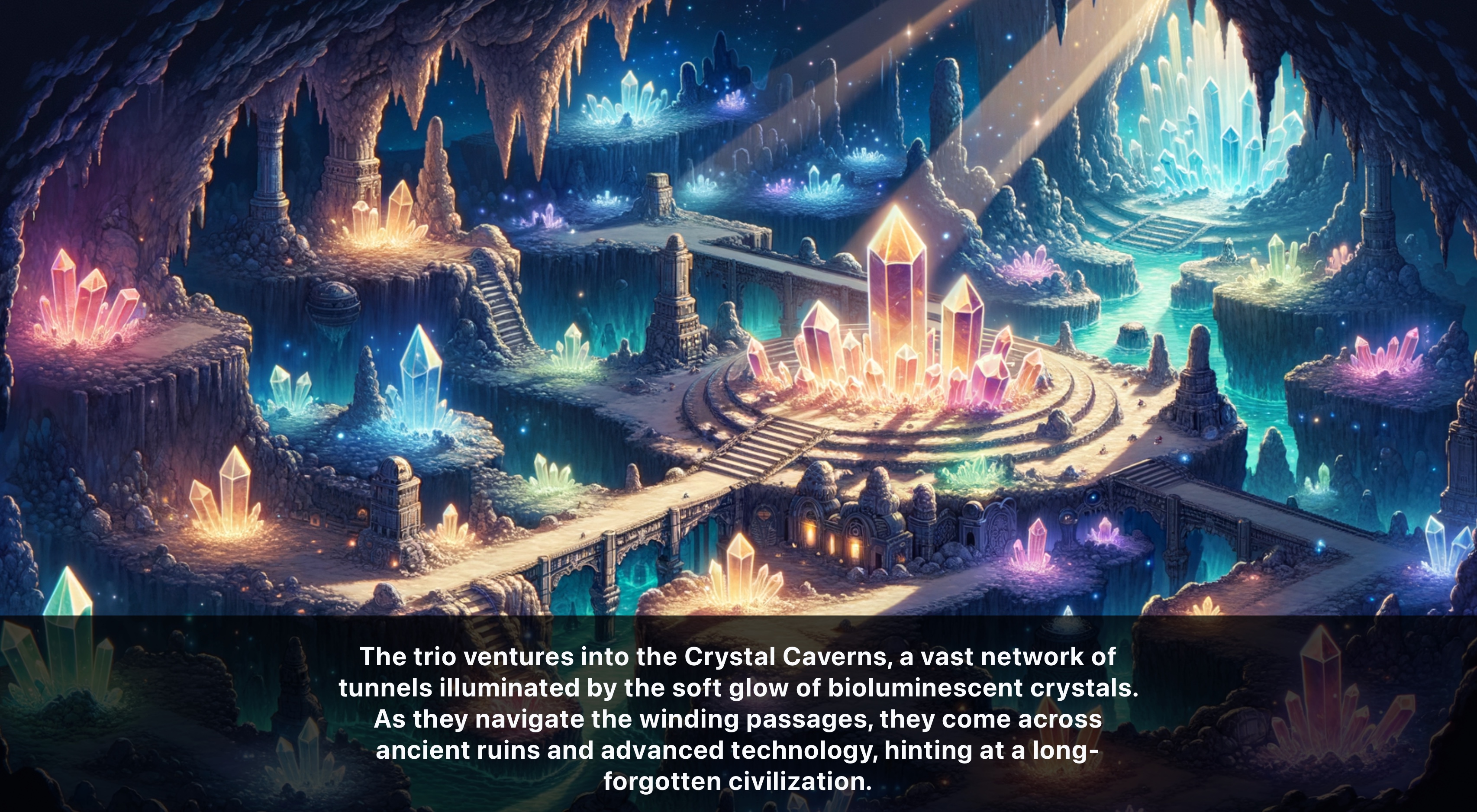}
     \end{subfigure}
    \begin{subfigure}[tbp]{\linewidth}
         \centering
         \includegraphics[width=\textwidth]{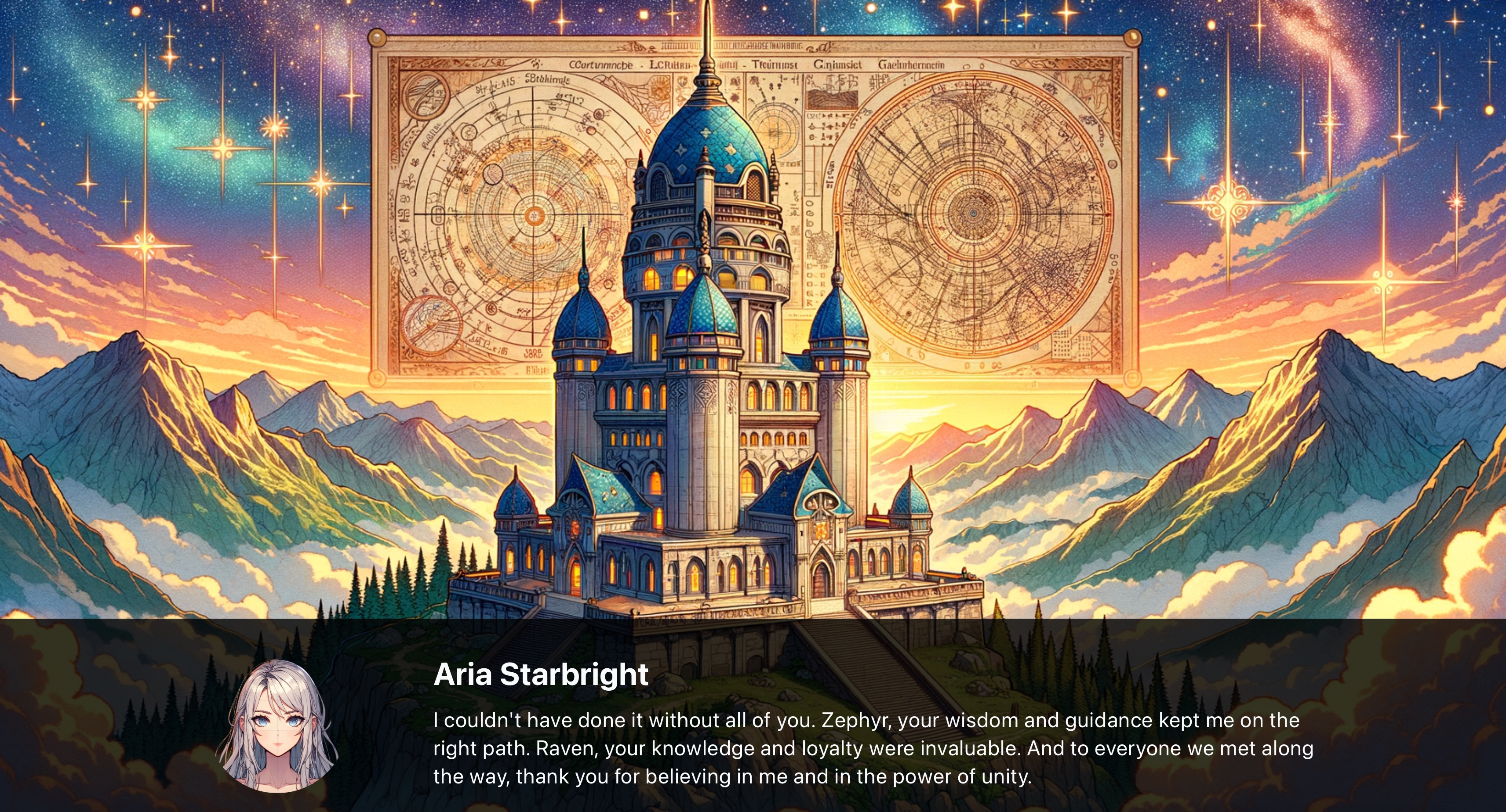}
     \end{subfigure}
     \caption{Additional screenshots of the best game generated using the DCP/P approach. Each screenshot is taken from various parts of the story during its progression.}
    \label{fig:additional_screenshot_proposed}
\end{figure}

In this subsection, we provide a synopsis of the best-performing game from each approach. This information should aid in understanding our qualitative analysis and discussion. \autoref{fig:synopsis_baseline} shows a synopsis of a game generated using the baseline approach, and \autoref{fig:synopsis_proposed} shows a synopsis of a game generated using the DCP/P approach. Additional screenshots of the games generated from the baseline and DCP/P approaches are available in Figures \ref{fig:additional_screenshot_baseline} and \ref{fig:additional_screenshot_proposed}, respectively.

\begin{figure}
    \centering
    \begin{mdframed}
    In a world where magic and technology intertwine, Zephyr Stormborn, a young adventurer, embarks on a quest to uncover the secrets of his past and the truth behind the ancient civilizations that once ruled the world. Guided by Aria Lightweaver, a powerful elven mage, and accompanied by Rook Ironheart, a dwarven inventor, Zephyr must navigate through enchanted forests, sunken cities, and floating islands to reach the Celestial Citadel, said to hold the key to his destiny. Along the way, he encounters Nyx Shadowcloak, a mysterious shapeshifter, and Orion Starhunter, a rival explorer, as he races to unravel the mysteries of The Rift and prevent a cataclysmic event that threatens to tear the world apart.
    \end{mdframed}
    \caption{A synopsis of the best-performing story generated using the baseline approach: \textbf{The Chronicles of Zephyr}.}
    \label{fig:synopsis_baseline}
\end{figure}

\begin{figure}
    \centering
    \begin{mdframed}
    In a world where magic and technology intertwine, Aria Starbright, a young adventurer, embarks on a quest to uncover the truth behind her mysterious past. Guided by an ancient prophecy and aided by a diverse group of companions, Aria must navigate through enchanted forests, abandoned spaceports, and celestial towers to piece together the fragments of her identity. As she unravels the secrets of her world, Aria discovers a sinister plot that threatens the delicate balance of the universe, and she must confront the enigmatic leader of the Celestial Guild to save her planet from destruction.
    \end{mdframed}
    \caption{A synopsis of the best-performing story generated using the DCP/P approach: \textbf{Celestial Odyssey}.}
    \label{fig:synopsis_proposed}
\end{figure}

\subsection{Additional Details of the Framework}\label{appendix:framework}

\subsubsection{Prompts}\label{appendix:framework:prompts}

We design various prompts for different purposes in interacting with generative models. There are a total of five prompt templates for an LLM that we designed for different purposes. Each prompt template utilizes a pattern of template \cite{white2023prompt} to specify JSON structured output template. This decision is made to aid in the automation process of output extraction by supporting systems. We also incorporate a JSON magic phrase in all prompt templates to enhance the reliability of generating a structured output. This phrase is adapted from a phrase used to improve the reliability of LLMs in generating a Markdown code block for level generation tasks \cite{abdullah2024the}.

The first prompt template is a \textit{plot prompt}. The plot prompt instructs the LLM to generate story data, including information such as title, synopsis, and chapter synopses, given the specified arguments. The second and third prompts are used to generate a \textit{story chunk} before the ending of a chapter or game. The differences between these two prompts are whether we provide information on a selected choice or not. However, they are largely the same, providing information about the current chapter out of all chapters, current chapter synopsis, and selected choice in the case of the prompt used to generate a story chunk based on a story choice. It is crucial to note that our prompt instructs an LLM to generate a story that progresses by multiple dialogues by a character or a narrator as an array of complex objects. The LLM is also instructed to generate a summary of the story so far. This approach ensures that the LLM is grounded in the current story before generating a new part.

Finally, the fourth and fifth prompts are used to generate a story chunk that serves as a chapter ending chunk or game ending chunk (leaf node). Both prompts share similarity in the instruction. However, in the case of the chapter ending prompt, we also provide the next chapter synopsis, while in the game ending prompt, we provide information on the selected game ending.

Aside from these prompt templates for the LLM, we also craft two additional prompt templates for image generative models used to generate assets for characters and scenes. For character assets, we instruct a model to generate a profile image of a 2D character based on their basic profile information, i.e., name, age, gender, species, and their physical appearance. On the other hand, the scene prompt is based on the scene name, location, and description.

\subsubsection{Supporting Systems}\label{appendix:framework:support}

Our framework utilizes Neo4J\footnote{\url{https://neo4j.com}}, a graph database, to facilitate efficient data management and retrieval. By leveraging Neo4J's graph-based structures, we can manage interconnected data effectively, supporting the dynamic creation of story branches and decision paths. This capability enhances our platform's ability to adapt and scale, providing robust support for a wide range of interactive storytelling and content creation applications. As previously stated, we can structure a typical visual novel game's story as a graph. Therefore, a graph database is very suitable for storing this type of data.

Additionally, our framework includes an orchestration program with a retry mechanism. The retry mechanism enhances the reliability of content generation by attempting to generate content multiple times until success or exhaustion of retries. LLMs are not perfect due to their inherent stochastic nature; therefore, the generation of malformed structured output is unavoidable. The mechanism ensures robustness against transient errors or validation issues encountered during the generation process. Each attempt involves validating the generated story chunk and associated choices, logging validation errors or exceptions, and incrementing the retry counter. If all attempts fail to produce valid content, the mechanism logs detailed error messages and exits the script.

We also leverage DALL-E 3 \cite{betker2023improving}, an advanced image generative model from OpenAI. We employ this model to create diverse visual assets through customized prompts, designed to generate high-quality images while avoiding undesirable attributes. In addition, our framework utilizes BriaRMBG-1.4\footnote{\url{https://huggingface.co/briaai/RMBG-1.4}}, an advanced background removal model that uses deep learning techniques to seamlessly remove backgrounds from images. Integrated into our system, BriaRMBG-1.4 processes character and scene images for background removal while preserving image quality. This step enhances the immersive experience of the game.

To enhance variability in interactive storytelling and content creation, our framework incorporates randomization. Within our framework, we have an LLM decide almost everything, including determining main characters, possible locations, available choices, the impact of each choice on the story's progression, and whether to use a narrator or character to advance the plot. Additionally, game endings are also randomized by an LLM to inject unpredictability and increase engagement in the narrative experience. This approach not only enriches each interaction by offering unique decision-making scenarios but also encourages variability in each generated game.

\subsubsection{Web Game Interface}\label{appendix:framework:ui}

We develop our own web game interface that communicates with the graph database to display any games generated by the LLM. Our game interface allows users to play the visual novel with the ability to choose any available options. While there are many potential features that exist in a typical visual novel game, we keep this initial version of the game interface minimal. We expect that the game interface will be improved over time as the algorithm evolves with new features.

We decide to use web technology as our main platform because it requires no installation on the player's part and enables them to experience the game easily. Web technology also supports responsive design, allowing us to provide an experience on mobile devices as well. We also implement a fallback mechanism to display a default image for a character or a scene. Since LLMs are not perfect and there is a chance that a model might select a non-existent character or scene to display, this mechanism helps provide a smoother gameplay experience.

\end{document}